\documentclass[final]{cvpr}

\usepackage{times}
\usepackage{epsfig}
\usepackage{graphicx}
\usepackage{amsmath}
\usepackage{amssymb}

\usepackage{bm}
\usepackage{enumitem}
\usepackage{multirow}
\usepackage{color}
\newcommand{\minisection}[1]{\vspace{0.04in} \noindent {\bf #1}\ \ }

\usepackage[ruled,vlined]{algorithm2e}
\usepackage{amsmath}
\usepackage{amssymb}
\usepackage{pifont}
\usepackage{subcaption}
\newcommand{\cmark}{\ding{51}}
\newcommand{\xmark}{\ding{55}}

\newcommand{\alert}[1]{{\color{black} #1}}

\newcommand{\tabincell}[2]{\begin{tabular}{@{}#1@{}}#2\end{tabular}} 

\usepackage[pagebackref=true,breaklinks=true,colorlinks,bookmarks=false]{hyperref}

\def\cvprPaperID{49} 
\def\confYear{CVPR 2021}

\begin{document}

\title{ACAE-REMIND for Online Continual Learning with \\Compressed Feature Replay}

\author{Kai Wang$^{1}$, Luis Herranz$^{1}$, Joost van de Weijer$^1$\\

$^{1}$ Computer Vision Center, Universitat Autònoma de Barcelona, Barcelona, Spain\\
{\tt\small \{kwang,lherranz,joost\}@cvc.uab.es}
}

\maketitle

\begin{abstract}
Online continual learning aims to learn from a non-IID stream of data from a number of different tasks, where the learner is only allowed to consider data once. Methods are typically allowed to use a limited buffer to store some of the images in the stream. Recently, it was found that feature replay, where an intermediate layer representation of the image is stored (or generated) leads to superior results than image replay, while requiring less memory. Quantized exemplars can further reduce the memory usage. However, a drawback of these methods is that they use a fixed (or very intransigent) backbone network. 
\alert{This significantly limits the learning of  representations that can discriminate between all tasks.} 
\alert{
To address this problem, we propose an auxiliary classifier auto-encoder (ACAE) module for feature replay at intermediate layers with high compression rates.}
The reduced memory footprint per image allows us to save more exemplars for replay. 
In our experiments, we conduct task-agnostic evaluation under online continual learning setting and get state-of-the-art performance on ImageNet-Subset, CIFAR100 and CIFAR10 dataset.
\end{abstract}


\section{Introduction}
The vast majority of deep learning papers consider that all training data is available jointly, and the learner can process the data several times (epochs) to learn the optimal parameters to solve the task at hand. However, in many real-world scenarios, this would not be possible, and the learner has only access to data of a single task at the time, before proceeding to learn a new task.
\alert{
This scenario refers to \textit{continual learning} (or \textit{incremental learning, lifelong learning}). The main challenge in this scenario is to learn from the current data while preventing forgetting the knowledge of previous tasks. With a naive finetuning approach the model will suffer a drastic drop in performance on previous tasks because the model aims to be optimal for the current tasks, and ignores performance on previous tasks.
This phenomenon is known as \textit{catastrophic forgetting~\cite{kirkpatrick2017overcoming,mccloskey1989catastrophic}.}
}
The field of continual learning studies methods that prevent forgetting~\cite{douillard2020podnet,hayes2019remind,hou2019learning,rebuffi2017icarl,wu2019large,masana2020ternary}.

A challenging setting in continual learning, yet common in practical application, is \textit{online continual learning} of non-iid data streams~\cite{aljundi2019online,chaudhry2018efficient,lopez2017gradient}. In the online setting, each image can only be observed once during model optimization (except exemplars in storage). These applications mainly exist in resource constrained devices, such as mobile phones, robots and other smart devices. The majority of methods in continual learning, known as batch incremental learning methods, allow for several cycles (epochs) over the data~\cite{de2019continual}. These methods cannot operate in the challenging online continual learning setting.  Moreover they take longer to train. In this paper, we focus on online continual learning.

Among the approaches to address catastrophic forgetting, some of the best performing ones are rehearsal-based~\cite{rebuffi2017icarl,shin2017continual,hayes2019remind}. Several methods save a small set of exemplar images of previous classes~\cite{rebuffi2017icarl,chaudhry2018efficient,hou2019learning,wu2019large}. Retrieving them during future training is a straightforward way to prevent forgetting. For example, GEM~\cite{lopez2017gradient}, A-GEM~\cite{chaudhry2018efficient} and MIR~\cite{aljundi2019online}, which address online continual learning, belong to this type. However, this strategy leads to increased memory usage and the problem of training from imbalanced data (between previous tasks and the current task). An alternative is to generate images via generative models (e.g. GANs)~\cite{shin2017continual,wu2018memory}. However, image generation is still a difficult problem in computer vision and requires complex generative models, which would also need to be continually learned, making this method not practical for complex datasets. 

To circumvent the difficulties of image replay, recent work has focused on feature replay~\cite{hayes2019remind,liu2020generative}. In~\cite{liu2020generative}  a generator is trained to replay \alert{compact feature representations of the images} (after the last average pooling layer of a ResNet-18). In addition, a distillation loss was applied to prevent forgetting of the feature extractor. Instead of generating features, it was also observed by Hayes \textit{et al.}~\cite{hayes2019remind} that saving them as exemplars is very efficient, since it required less memory per image. To even further reduce the memory requirements, the REMIND method~\cite{hayes2019remind} also applies product quantization~\cite{jegou2010product}. 
\alert{
This allows them to save up to 1M compressed feature exemplars, instead of 20K exemplar images saved traditionally, in the same memory buffer.
}
A major drawback of these feature replay methods~\cite{liu2020generative,hayes2019remind} is that they either allow for very little training~\cite{liu2020generative} or no training at all~\cite{hayes2019remind}  of the backbone feature extractor (located before the replay layer). As a consequence, if this backbone is not yet optimally trained for future tasks, the performance is sub-optimal. 

To address the limitation of feature replay, we propose ACAE-REMIND, an auxiliary classifier auto-encoder (ACAE) that allows for compressed feature replay (as in REMIND) at intermediate layers of the network. This contrasts with current feature replay methods that focus on replaying the features in the last layers. The principal advantage of our method is that we can jointly train all layers after the replay layer. This addresses an important problem of feature replay methods, namely the reduced performance because of a large fixed backbone network. Instead of only the 4M parameters that are trained in REMIND when replaying at block 4 of a ResNet-18, we allow to train 9M parameters jointly when replaying on block 3. This leads to feature representations that are more discriminative  between the classes of current and previous tasks. 
We evaluate our method in the challenging, yet more realistic, \textit{online continual learning} setting. From experiments under multiple settings and datasets, we observe state-of-the-art performance in many-task evaluations and competitive in few-task settings.

\section{Related work}

\subsection{Continual learning}
Continual learning methods can be categorized into three types which we will shortly comment. For a more elaborate overview see the following surveys~\cite{de2019continual,masana2020class}).

\minisection{Regularization-based methods.} The first group of techniques is based on regularization. These methods add a regularization term to the loss function which impedes changes to the parameters deemed relevant to previous tasks. The difference depends on how to compute the estimation. From these differences, these methods can be further divided into data-focused~\cite{li2017learning} and prior-focused~\cite{kirkpatrick2017overcoming}. Data-focused methods use knowledge distillation from previously learned models. Prior-focused methods estimate the importance of model parameters as a prior for the new model. \alert{However, it has been shown that data-focused methods are vulnerable to domain shifts~\cite{aljundi2017expert} and prior-focused methods might be not sufficient to restrict the optimization process to keep acceptable performances on previous tasks~\cite{farquhar2018towards}.}

\minisection{Parameter isolation methods.} This family focuses on allocating different model parameters to each task. These models begin with a simplified architecture and updated incrementally with new neurons or network layers in order to allocate additional capacity for new tasks. In Piggyback/PackNet~\cite{mallya2018piggyback,mallya2018packnet}, the model learns a separate mask on the weights for each task, whereas in HAT~\cite{serra2018overcoming}, masks are applied to the activations. \alert{This method is further developed to the case were no forgetting is allowed in~\cite{masana2020ternary}. In general, this branch is restricted to the task-aware (task incremental) setting. Thus, they are more suitable for learning a long sequence of tasks when a task oracle is present and there is no constraint over model capacities.
}

\minisection{Replay methods.} This type of methods prevent forgetting by including data (real or synthetic) from previous tasks, stored either in an episodic memory or via a generative model. There are two main strategies: exemplar rehearsal~\cite{rebuffi2017icarl,chaudhry2018efficient,hou2019learning,wu2019large} and pseudo-rehearsal~\cite{shin2017continual,wu2018memory}. The former store a small amount of training samples (also called exemplars) from previous tasks. The latter use generative models learned from previous data distributions to synthesize data. 

One of the main drawbacks of exemplar replay is the high memory usage required to store exemplars of previous tasks. 
REMIND~\cite{hayes2019remind} addresses this drawback, instead of saving original data, it saves compressed latent representation of intermediate layer features via product quantization~\cite{jegou2010product}. This is a more efficient usage of memory and computation. However, due to restriction that the backbone is fixed, the majority of feature extraction modules cannot be adopted to later tasks. Therefore, this model has a strong bias towards the first task. Recently, GDumb~\cite{prabhu2020gdumb} proposes training a model only from exemplars. The main idea is to balance the sample reservoir in a selection stage, then the model is learnt from scratch on this balanced set. While not designed for any specific continual learning settings, it achieves excellent performance on many. It reveals that sample balancing is crucial for rehearsal-based continual learning methods. While saving images is always expensive compared to saving features, this point is also mentioned in paper on feature adaption~\cite{iscen2020memory}.

\subsection{Auto-encoders and product quantization}
Auto-encoders~\cite{kramer1991nonlinear} learn representations in an unsupervised way by encouraging the model to reconstruct the input data. An encoder projects the high-dimensional input to a low-dimensional space, and the decoder tries to project back to the original space minimizing the reconstruction error. Product quantization~\cite{jegou2010product} is an effective quantization method that performs a decomposition of a high-dimensional space into the Cartesian product of a series of subspaces, and quantizes them separately.

In our model, we propose an auto-encoder with an auxiliary classifier (ACAE) to force the reconstructions not only to remain close to original inputs but also keeping the classification characteristics. 
By combining ACAE with Product Quantization (PQ), the feature spaces are decomposed from high dimension to low dimension, from float numbers to integer indexes, which leads to better compression and therefore allows to save more exemplars.

\begin{figure}[t]
\begin{center}

\includegraphics[width=1\linewidth]{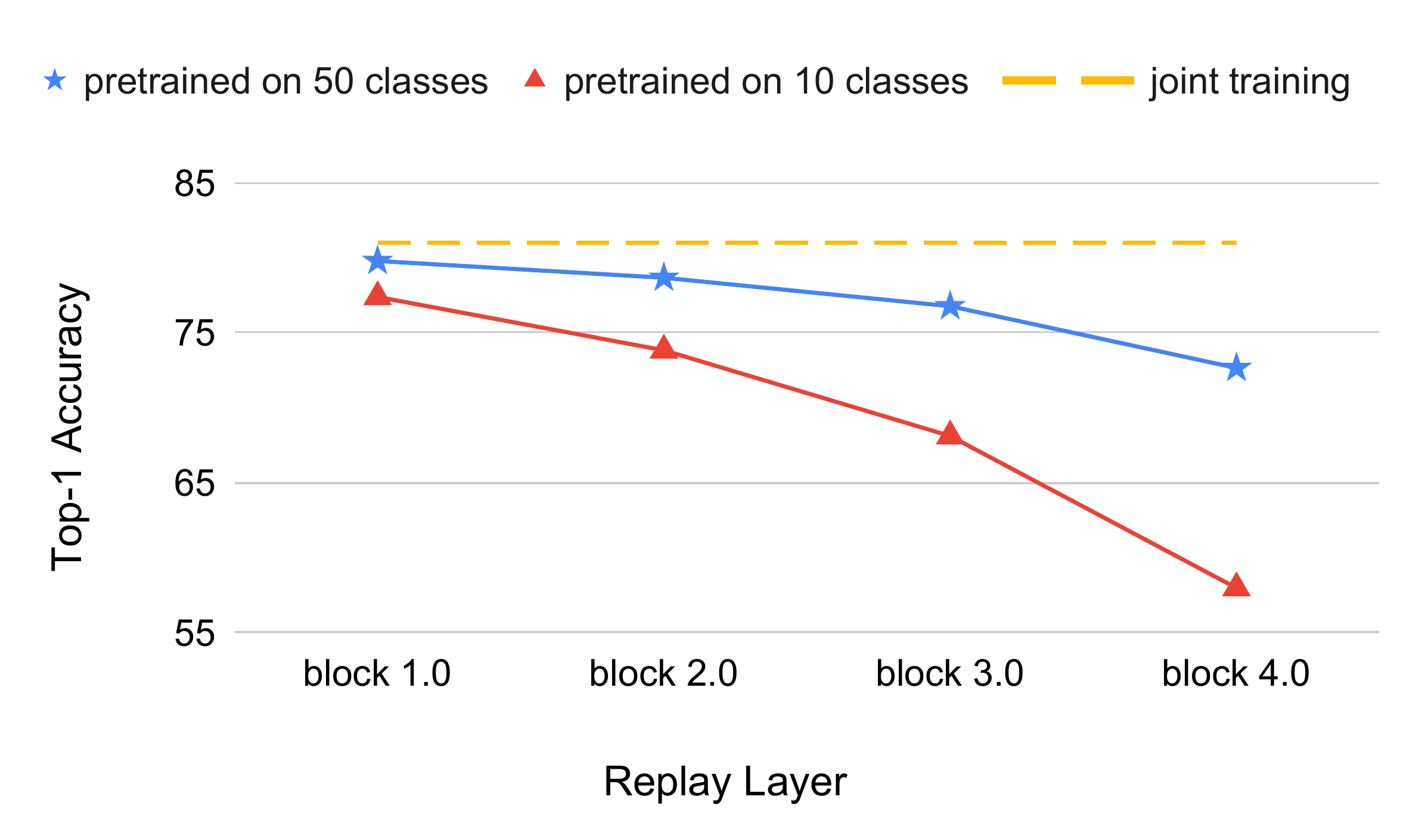}
\end{center}
   \caption{Drop in performance due to frozen backbones (Joint training: 81.0)}
\label{fig:chenshen}
\end{figure}

\section{Compressed Feature Replay}
\subsection{Feature replay location}

Pseudo-rehearsal methods~\cite{shin2017continual,wu2018memory} are limited by the performance of generative models to generate high-quality images. As a results these methods perform poorly on more complex real-world datasets. To address this limitation, Liu \textit{et al.}~\cite{liu2020generative} proposed generative feature replay (GFR) to generate features of an intermediate layer. In their proposal, features before the classifier are generated by a conditional GAN learned in a continual fashion.

REMIND~\cite{hayes2019remind} observes that storing features is much more efficient than storing images. They operate on the same block as GFR. With the help of PQ~\cite{jegou2010product} the features are further compressed in REMIND. The features from block 4.0 of ResNet-18 are approximated by a number of codebooks and indexed feature maps. By this means, the floating point values of feature representations are replaced by integer index numbers. This allows them to save 50$\times$ more feature exemplars than image exemplars in the same memory space and obtain excellent results for online continual learning. 

As noted in the introduction, one of the main drawbacks of REMIND (and feature replay in general) is that these methods freeze the backbone feature extractor (i.e the layers before the feature replay) after training the first task. They only train the layers that come after the feature replay layer for the remaining tasks. Depending on the continual learning scenario this could lead to a significant drop in performance because of a suboptimal backbone network.

To better understand the impact of freezing the backbone network after the first task we perform an experiment on the ImageNet-Subset dataset~\cite{imagenet_cvpr09} with ResNet-18 as the backbone. We consider two scenarios, one with the first task containing 50 classes and the remaining 50 classes divided into five more tasks. In the second scenario, we evenly divide the classes over 10 tasks (each with 10 classes). Clearly, the second scenario is more challenging for REMIND, because now the backbone network can only be trained on the 10 classes of the first task. In Fig.~\ref{fig:chenshen} we can see the drop in performance which is caused by freezing the backbone network as a function of the position of the feature replay. The performance of the different backbone networks is computed in the following way: we first train the first task and fix the backbone network, then we jointly train the remaining layers on all training data of all tasks (this can be seen as the upper bound for this continual learning setting, i.e. joint training with the backbone frozen). As can be seen, the drop in performance is significant (by comparing the difference of the blue and red with the yellow line), dropping 8.36\% in the first scenario, and 23.04\% in the second scenario when replaying the features of block 4.0. As can be seen the drop diminishes considerably by performing the replay at earlier layers. The reason why REMIND chooses to replay at block 4.0 is because the proposed technique does not scale well to lower positions in the network. This is explicitly discussed and they mention that the quantized features would be too large and would significantly increase storage requirements\footnote{See Supplementary material S2~\cite{hayes2019remind}.}.

To overcome the limitations of feature replay, our proposed ACAE-REMIND model aims to apply replay on an intermediate blocks.   
To reconstruct features in intermediate layers, we introduce a stronger compression module, which achieves dimension reduction, and feature approximation while maintaining the classification characteristics of the replayed features. The method is an extension of REMIND and is based on an Auto-Encoder with Auxiliary Classifier (ACAE). We can perform joint training on all layers after the replay layer (and not only the last block as in REMIND). This alleviates the drawback of fixing the backbone neural network. A comparison among the discussed feature replay methods is in Table~\ref{tab:replay_method_comparison}.

\begin{table}
\begin{center}
\caption{\label{tab:replay_method_comparison}Comparison of replay methods.}

\scalebox{0.85}{
\begin{tabular}{|c|c|c|c|}
\hline
method name & replay layer & compression & online\\
\hline\hline

GFR &  last block & GAN& \xmark \\
REMIND & last block  & PQ  & \cmark\\
Ours &  interm. block & ACAE+PQ & \cmark\\

\hline
\end{tabular}
}
\end{center}

\end{table}

\begin{figure}[t]
\begin{center}

\begin{subfigure}{0.5\textwidth}
\centering
  \includegraphics[width=1\linewidth]{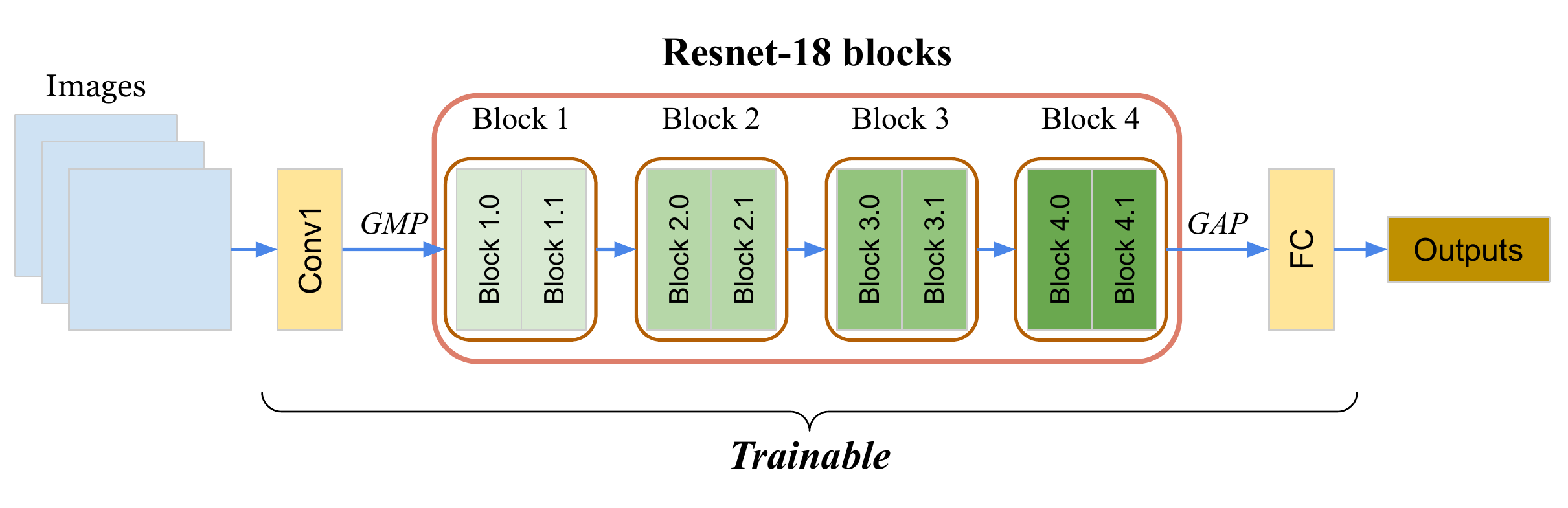}
  \subcaption{Step 1: Classification model (Resnet-18) training} \label{fig:1a}
\end{subfigure}

\begin{subfigure}{0.5\textwidth}
\centering
  \includegraphics[width=1\linewidth]{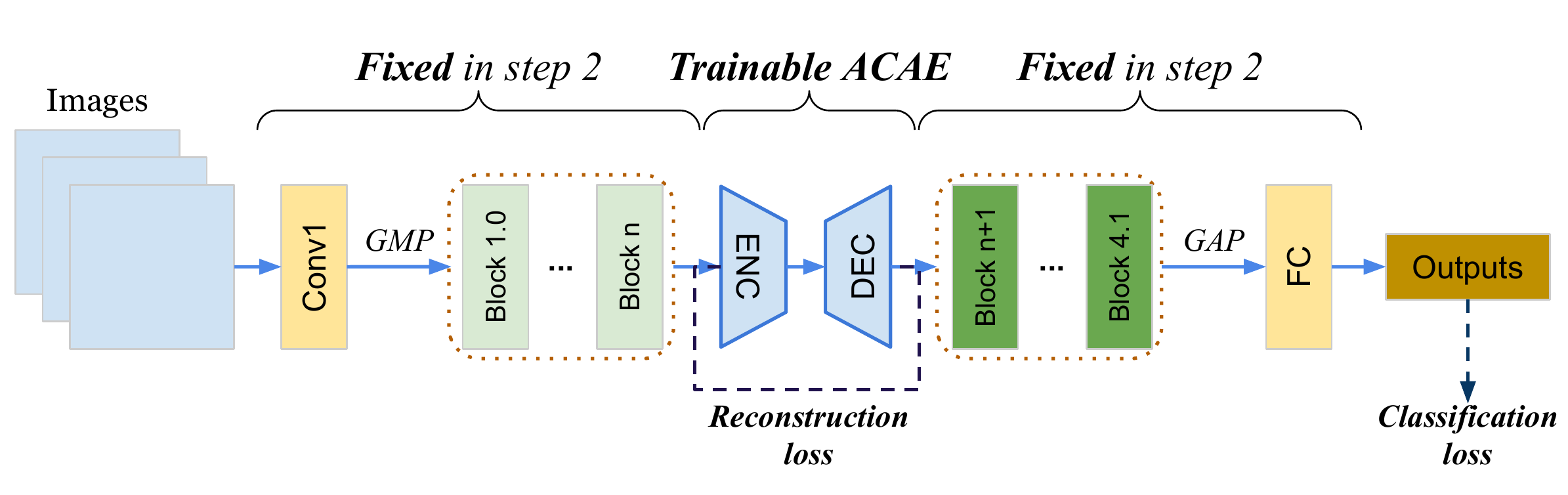}
  \subcaption{Step 2: ACAE (Auxiliary classifier auto-encoder) training} \label{fig:1b}
\end{subfigure}

\begin{subfigure}{0.5\textwidth}
\centering
  \includegraphics[width=1\linewidth]{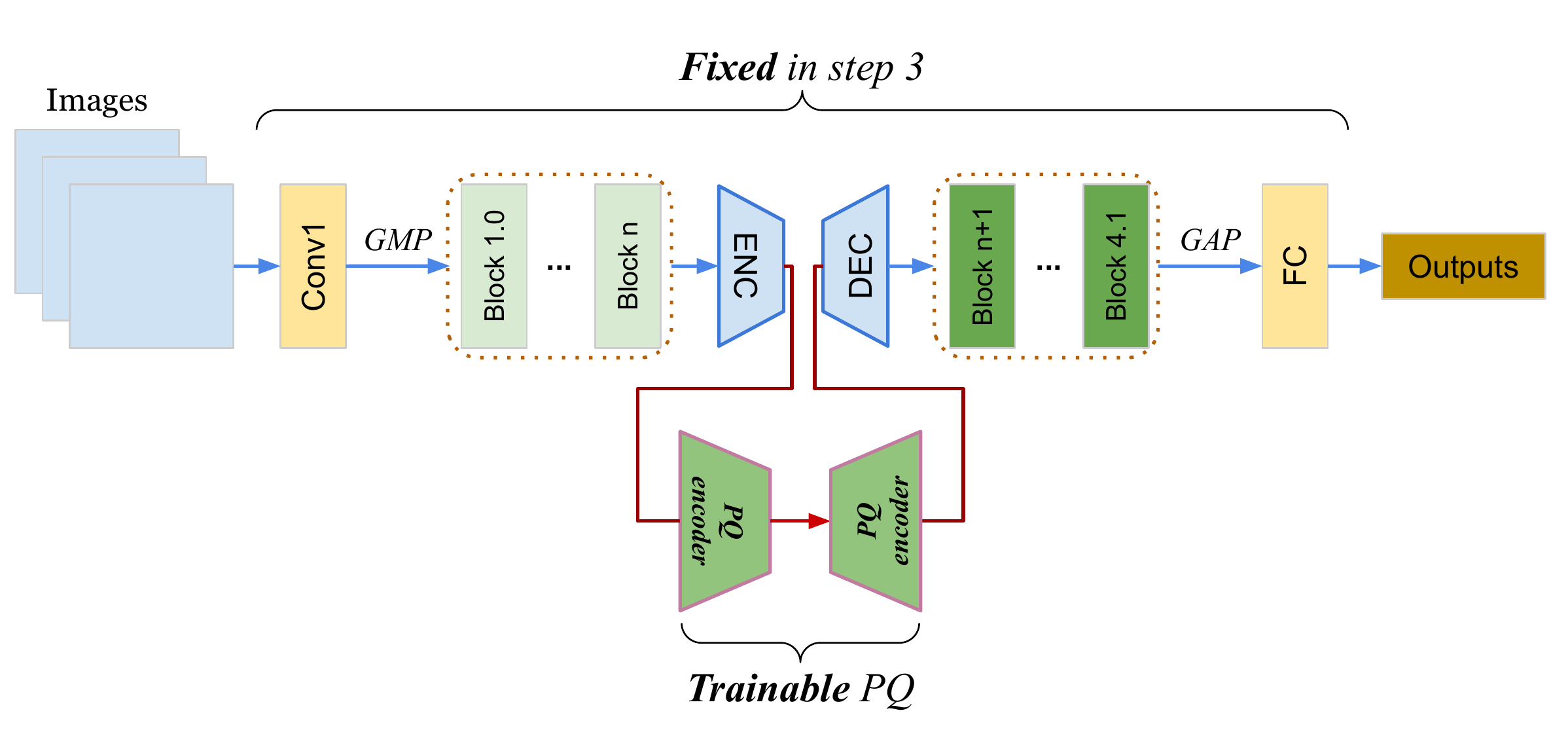}
  \subcaption{Step 3: PQ (Product Quantization) training} \label{fig:1c}
\end{subfigure}

\end{center}
   \caption{Overview of the initialization stage (trained on first task).}
\label{fig:initialization}
\end{figure}

\begin{figure}[t]
\begin{center}

\includegraphics[width=1.08\linewidth]{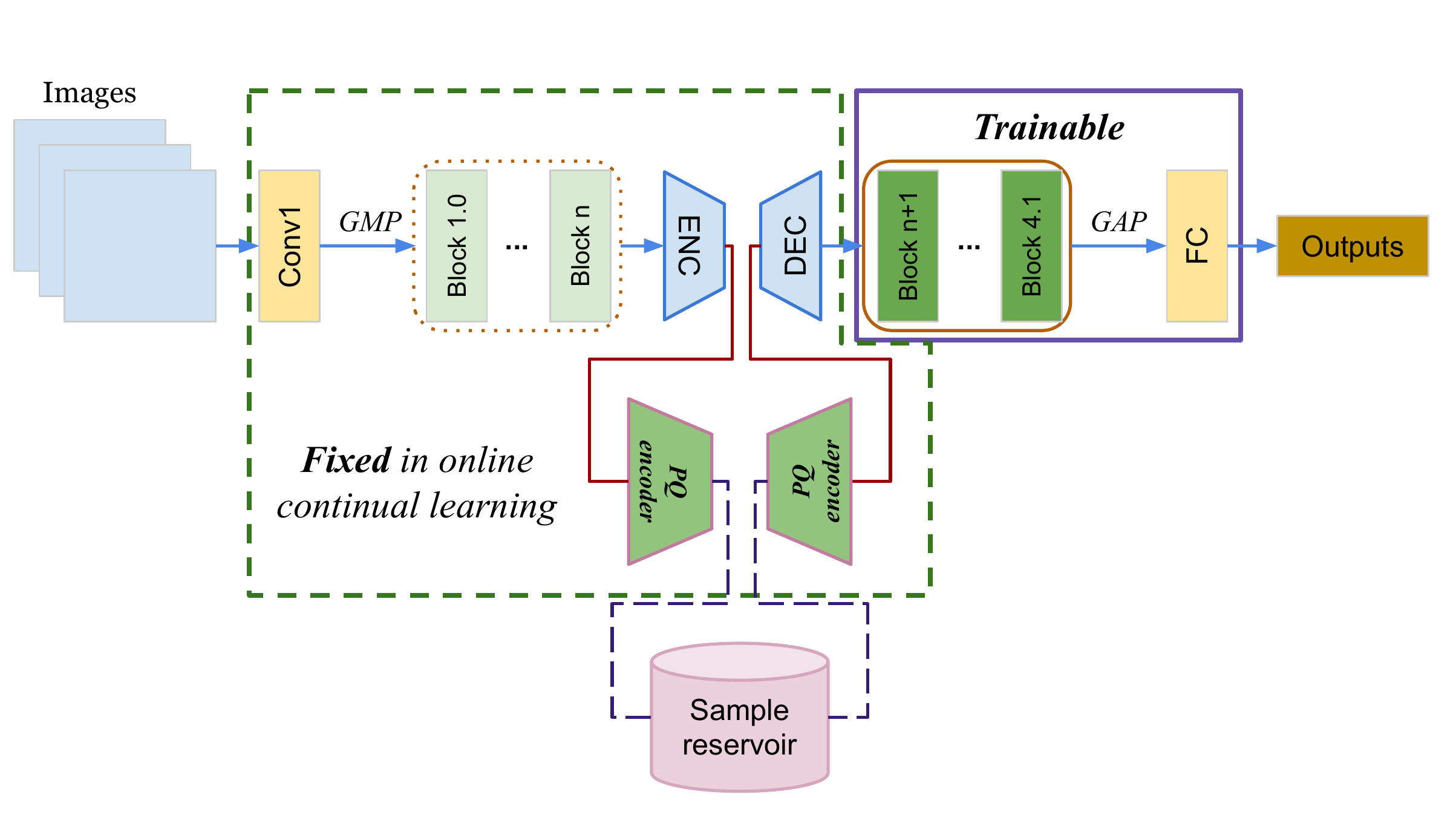}
\end{center}
  \caption{Overview of our online continual learning phase (task $t=2,...,T$).}
\label{fig:CL}
\end{figure}

\subsection{Online continual learning setting}

Online continual learning is a subarea of incremental learning, where the algorithm is only allowed to make a single pass through the data of each task. It is more related to real-life and real-time applications since data comes in sequential streams, and they are not allowed to use the same sample more than one time (unless stick to buffer) in the whole learning process. 

Suppose we have a data stream of triplets $\left\{\left ( x_{t}^{i},y_{t}^{i},t \right )\right\}_i $, where $x_{t}^{i}$ is the $i$-th input, $y_{t}^{i}$ is the corresponding label and $t$ is the task identifier ($t\in \text{\Large$ \tau $} = \left \{ 1,...,T \right \}$ ). Each input-label pair is an identical and independent sample drawn from an unknown distribution $P_{t}\left ( X,Y \right )$ of task $t$. We consider the number of tasks $T$ is unknown and the tasks are coming \alert{sequentially} as $t=1,...,T$. We also assume that data among tasks are disjoint. At inference time, the task-id $t$ is unknown at all time; also referred to as task-agnostic inference. Under this assumption, the resulting model $f\left ( x;\Theta \right )$, parameterized by $\Theta$, is optimized to minimize a predefined loss $l\left ( x,y;\Theta  \right )$ over new sequential input samples $\left (  x_{t}^{i}, y_{t}^{i} \right )$ from current data stream $t$. And at the same time, the performance on previous tasks should not decrease.

\subsection{ACAE-REMIND for compressed feature replay}

The ACAE-REMIND model is designed for online task-agnostic continual learning in a memory efficient way. As explained before, we aim to execute feature replay on an intermediate layer. Since all layers after the feature replay can be jointly trained, this strategy can potentially lead to improved performance. Because the distribution in lower layers is more complex, we propose an improved compression mechanism based on an ACAE module. The whole training procedure can be roughly divided into: 1) initialization stage (in Fig.~\ref{fig:initialization}) of the classification model, ACAE and PQ modules, 2) online continual learning stage (in Fig.~\ref{fig:CL}).

\subsubsection{Initialization}
During initialization, the classification model, ACAE and PQ are trained sequentially with data $\left (  x_{1}^{i}, y_{1}^{i} \right )$ from the first task $t=1$. In the first step, the whole classification model is optimized in an offline way. This step aims to learn a robust pretrained model for future tasks (similar as in REMIND). The parameters $\Theta$ are  updated by minimizing the cross-entropy loss:

\begin{equation}
\label{celoss2}
 minimize_{\Theta} \mathcal{L}_{CE} (y_{1}^{i},\hat y_{1}^{i}  )  =- y_{1}^{i} \cdot  log\; \hat{y}_{1}^{i}
\end{equation}
\alert{
where the prediction is given by $\hat{y}_{1}^{i}=f\left ( x_{1}^{i};\Theta \right )$.}

Secondly, the ACAE module is inserted into the layer where we will replay the features. We denote the layers before and after the replay layer as $g\left ( x;\Theta_{1} \right )$ and $h\left ( z;\Theta_{2} \right )$, where $\Theta$ is the union of $\Theta_{1}$ plus $\Theta_{2}$ and $z=g\left ( x;\Theta_{1} \right )$ in step 1. The encoder and decoder of ACAE are denoted as $u=D_{enc}\left ( z;\Gamma \right )$ and $\hat{z}=D_{dec}\left ( u;\Pi \right )$. The ACAE is trained with an auxiliary classification loss and auto-encoder reconstruction MSE(mean square error) loss on the same data stream $\left (  x_{1}^{i}, y_{1}^{i} \right )$.

Then parameters $\Gamma,\Pi$ are computed by minimizing the ACAE loss:

\begin{equation}
\label{eq:acae_loss}
minimize_{\Gamma,\Pi}\ \mathcal{L}_{ACAE}(y_{1}^{i},x_{1}^{i} )    = \mathcal{L}_{CE}(y_{1}^{i},\hat y_{1}^{i} )    +  \left \|z_{1}^{i}-\hat{z}_{1}^{i}  \right \|_{2}
\end{equation}

where

\begin{equation}
\begin{split}
z_{1}^{i} &= g\left ( x_{1}^{i};\Theta_{1} \right )\\
\hat{z}_{1}^{i}&= D_{dec}\left (D_{enc}\left (z_{1}^{i};\Gamma  \right ), \Pi  \right )\\
\hat{y}_{1}^{i}&= h\left ( \hat{z}_{1}^{i};\Theta_{2} \right ).
\end{split}
\end{equation}

After that, the last step is to train a PQ encoder-decoder pair $P_{enc}\left ( u; \Upsilon  \right )$ and $P_{dec}\left ( v; \Psi \right )$ to approximate latent representations extracted from ACAE's encoder. Here $\Upsilon,\Psi$ are learnt from the object function of PQ MSE loss:

\begin{equation}
\begin{split}
minimize_{\Upsilon,\Psi} &  \  \mathcal{L}_{PQ}\left(x_1^i\right) = \left \|u_{1}^{i}-\hat{u}_{1}^{i}  \right \|_{2}\\
z_{1}^{i}=&(g\left ( x_{1}^{i};\Theta_{1} \right )\\
u_{1}^{i}=& D_{enc} \left (z_{1}^{i};\Gamma \right )\\
\hat{u}_{1}^{i}=&P_{dec}\left (P_{enc}\left (u_{1}^{i};\Upsilon \right ) , \Psi  \right )
\end{split}
\end{equation}

\subsubsection{Online continual learning}
In online continual learning, only layers after the ACAE decoder can freely adjust to new tasks (the parameters in $\Theta_{2}$), other modules (including parameters $\Theta_{1}, \Gamma, \Pi,\Upsilon, \Psi$) are all fixed during training. The new coming images from the data stream are passed through lower layers, the ACAE encoder and the PQ encoder to get their latent representations $v_{t}^{i}$ with corresponding integer indexes. This is computed as:

\begin{equation}
\label{daeloss1}
v_{t}^{i}=P_{enc}\left (D_{enc}\left (g\left ( x_{t}^{i};\Theta_{1} \right );\Gamma \right ) , \Upsilon  \right )
\end{equation}

Then its representation is mixed with randomly selected $\mathcal{N}$ previous samples $v_{\hat{t}}^{j} \ (\hat{t}< t)$ from the reservoir to reconstruct features via the PQ decoder and the ACAE decoder. Those features will be taken to optimize the trainable parameters $\Theta_{2}$ with the cross-entropy loss $\mathcal{L}_{CE}\left ({y}_{\bar{t}}^{i},\hat{y}_{\bar{t}}^{i} \right )$ and $\hat{y}_{\bar{t}}^{i}$ is formed as:

\begin{equation}
\label{daeloss2}
\hat{y}_{\bar{t}}^{i}=h\left (D_{dec}\left (P_{dec}\left ( v_{\bar{t}}^{i};\Psi \right ) ,\Pi  \right )  , \Theta_{2} \right ), \ \bar{t} \leq t
\end{equation}

\minisection{Reservoir sampling}
After optimization, the new representation will be stored in the reservoir memory. If the reservoir is full, we randomly select a sample to pop up from one of the classes with most samples in the reservoir. 

\section{Experiments}
\subsection{Experimental setup}
\minisection{Datasets.}Our evaluations are performed on three datasets: ImageNet-Subset~\cite{imagenet_cvpr09}, CIFAR100~\cite{krizhevsky2009learning} both with 100 classes\footnote{Samples are presented in a random yet fixed presentation order, as proposed in iCaRL~\cite{rebuffi2017icarl}, and adopted by others~\cite{douillard2020podnet,hou2019learning,prabhu2020gdumb,wu2019large}.}
and CIFAR10 with 10 classes. We use data augmentation during the initial training of the full model and the ACAE, but removed it to train PQ (we save the representation of the original image - without augmentation -) and during the online continual learning stage. For feature augmentation, we only randomly resize and crop reconstructed features in the online continual learning stage.

\minisection{Implementation details.} We use Resnet-18 as our classification network for ImageNet-Subset. For CIFAR10 and CIFAR100, we use adapted Resnet-18 and Resnet-32 respectively (using only 3 blocks instead of the original 4 blocks). During initialization, the backbone network is learned from scratch with SGD, then the ACAE is trained with Adam~\cite{kingma2014adam}. For PQ training, we use the implementation from the Facebook Faiss library~\cite{JDH17}. During the online continual learning stage we use SGD.

\minisection{Evaluation metrics.} We consider two widely used metrics: Average of top-1 accuracy over classes (AOC) up to the current task and \alert{top-1} accuracy after the last task (LAST).

\minisection{Experimental settings.}We will evaluate our method in five different settings. For the first three settings on ImageNet-Subset and CIFAR100, we use half of the classes as the first task and split the remaining into 5, 25 and 50 tasks with equal split (this setting is widely used~\cite{douillard2020podnet,hou2019learning,prabhu2020gdumb}). We also refer to these as the 5, 25 and 50 steps setting. 
The fourth setting \alert{is splitting} ImageNet-Subset into 10 tasks of the same size (this setting is used in~\cite{rajasegaran2019random,rebuffi2017icarl,wu2019large}). We compare with several methods: iCaRL~\cite{rebuffi2017icarl}, BiC~\cite{wu2019large}, UCIR~\cite{hou2019learning}, PODNet~\cite{douillard2020podnet}, GDumb~\cite{prabhu2020gdumb}, RPSnet~\cite{rajasegaran2019random} and REMIND~\cite{hayes2019remind}. We note that, except REMIND, the other methods are mainly designed for offline continual learning, which is a simpler setting compared with our online setting. 

The fifth setting is on CIFAR10, where we use the commonly used setting from GMED~\cite{jin2020gradient}, which divides CIFAR10 into 5 tasks equally. And we compare with online continual learning methods: AGEM~\cite{chaudhry2018efficient}, BGD~\cite{zeno2018task}, GEM~\cite{lopez2017gradient}, GSS-Greedy~\cite{aljundi2019gradient}, HAL~\cite{chaudhry2020using}, ER~\cite{rolnick2019experience}, MIR~\cite{aljundi2019online}, and GMED~\cite{jin2020gradient}.

\subsection{Results of online continual learning}

\begin{table}
\begin{center}
\caption{\label{tab:base50tasks5}
Comparison on Imagenet-Subset, we show the averages over classes (AOC) with 50 classes as the first task and 5/25/50 steps each with 10/2/1 classes. For REMIND and our method, we show the replay layer (block number) in brackets. The highest numbers in each row are highlighted.}

\scalebox{0.74}{
\begin{tabular}{|c|c|c|c|c|c|c|c|}
\hline
\multirow{2}{*}{\tabincell{c}{On- \\ line}} & \multicolumn{3}{c|}{Exemplar info} & \multirow{2}{*}{Methods} & \multicolumn{3}{c|}{AOC over various steps} \\
\cline{2-4}
\cline{6-8}
 & Num. & \tabincell{c}{Shape \\ (CHW)} & \tabincell{c}{Mem. \\ (MB)}  & & 5  & 25  & 50   \\

\hline
\multirow{6}{*}{\xmark } & \multirow{6}{*}{2K} & \multirow{6}{*}{\tabincell{c}{3\\224\\224 \\ (int)}} & \multirow{6}{*}{301} & iCaRL & 65.56 & 54.56 & 54.97\\
&&&& BiC  & 68.97 & 59.65 & 46.49 \\
&&&& UCIR(NME) &  69.07 & 60.81 & 55.44\\
&&&& UCIR(CNN) &  71.04 & 62.94 & 57.25\\
&&&& PODNet(CNN) &  \textbf{75.54} & 68.31& 62.08\\
&&&& GDumb  & -& -& 62.86 \\
\hline
\multirow{6}{*}{\cmark} & \multirow{6}{*}{130K} &  \multirow{6}{*}{\tabincell{c}{8\\7\\7 \\ (int)}} & \multirow{6}{*}{51} & REMIND(3.0)  & 70.58 & 67.93 & 67.35\\
&&&& REMIND(4.0)   & 71.02 & 70.50 & 70.14\\

&&&& Ours(1.0) &  {60.70} &  56.37 & 56.10 \\
&&&& Ours(2.0) &  {70.24}& 67.65 & 66.23 \\
&&&& Ours(3.0) &  \textbf{72.58} & \textbf{71.43} & \textbf{70.69}\\

\hline
\multirow{6}{*}{\cmark} & \multirow{6}{*}{130K} & \multirow{6}{*}{\tabincell{c}{32\\7\\7 \\ (int)}} & \multirow{6}{*}{204} & REMIND(3.0)  & 72.46 & -& -\\

&&&& REMIND(4.0) & 73.98 & -& -\\
&&&& Ours(1.0) &  \textbf{74.08}& -& - \\
&&&& $-\mathcal{L}_{CE}$  &  70.26 & -& -\\
&&&& Ours(2.0) &  {73.75} & -& -\\

&&&& Ours(3.0) &  {73.63}& -& - \\

\hline
\end{tabular}
}
\end{center}
\end{table}

\minisection{Few-task evaluation (5 steps setting).} We report the AOC metric on ImageNet-Subset with the 5 steps setting in Table~\ref{tab:base50tasks5}. Every time we have a new input image, we randomly sample $\mathcal{N}=50$ previous latent representations from the sample reservoir.
It can be seen that our method outperforms REMIND and that, especially for larger memory, we can obtain excellent results by replaying lower layers. For comparison, we have also computed results for block 3.0 for standard REMIND (going to lower blocks did further reduce performance). We observe that our method with 32 codebooks is only 1.46\% lower than the state-of-the-art offline PODNet method, and it well outperforms other methods. Even when we have only 8 codebooks, it is still better than all offline algorithms except PODNet.
Another interesting phenomenon is that, with 32 codebooks, we get an increase from block 3.0 to block 1.0, but this trend gets reversed with only 8 codebooks. The reason is that in the lower layers, the latent representations contain more information and thus require more codebooks to be represented. 

For CIFAR100 with 5 steps, the performance is shown in Table~\ref{tab:cifar100}.  Due to smaller image-size, the compression ratio is not as high as in ImageNet-Subset. In this case, offline continual learning (PODNet) outperforms the online settings by a larger margin (6.1\%) under the same memory allocation. It should be noted that PODNet runs for 160 epochs over the data whereas the online methods can only do one epoch. \alert{Also, among the online methods, our method performs worse than REMIND(3.0) when considering a memory of 6.4MB. This is because a first task with many classes and more data allows REMIND to also learn a high-quality backbone network. For the larger memory setting (12.8MB) our method performs comparable to REMIND(3.0).}

\begin{table}
\begin{center}
\caption{\label{tab:cifar100}Comparison on CIFAR100 dataset, we show the averages over classes (AOC) with 50 classes as the first task and 5/25/50 steps each with 10/2/1 classes. For REMIND and our method, we show the replay layer (block number) in brackets. The highest numbers in each row are highlighted.}

\scalebox{0.72}{
\begin{tabular}{|c|c|c|c|c|c|c|c|}
\hline
\multirow{2}{*}{\tabincell{c}{On- \\ line}} & \multicolumn{3}{c|}{Exemplar info} & \multirow{2}{*}{Methods} & \multicolumn{3}{c|}{AOC over various steps} \\
\cline{2-4}
\cline{6-8}
 & Num. & \tabincell{c}{Shape \\ (CHW)} & \tabincell{c}{Mem. \\ (MB)}  & & 5  & 25  & 50   \\

\hline
\multirow{7}{*}{\xmark } & \multirow{7}{*}{2000} & \multirow{7}{*}{\tabincell{c}{3\\32\\32 \\ (int)}} & \multirow{7}{*}{6.14} & iCaRL & 58.08 & 50.60 & 44.20\\
&&&& BiC  & 56.86 & 48.96 & 47.09 \\
&&&& UCIR (NME) &  63.63 & 56.82 & 48.57\\
&&&& UCIR (CNN) &  64.01 & 57.57 & 49.30\\
&&&& PODNet (CNN) &  \textbf{64.83} & 60.72& 57.98\\
&&&& PODNet (NME) &  64.48 & \textbf{62.72}& \textbf{61.40}\\
&&&& GDumb & -& -& 58.40 \\

\hline
\multirow{4}{*}{\cmark} & \multirow{4}{*}{25000} &  \multirow{4}{*}{\tabincell{c}{4\\8\\8 \\ (int)}} & \multirow{4}{*}{6.40} & REMIND(2.0)  & 55.79 & 58.25 & 57.93\\
&&&& REMIND(3.0) & 58.71 & 59.26 & 58.99\\
&&&& Ours(1.0) & 53.38 & 53.74 & 54.25\\
&&&& Ours(2.0) & 57.57 & 59.28 & 59.50\\

\hline
\multirow{5}{*}{\cmark} & \multirow{5}{*}{50000} &  \multirow{5}{*}{\tabincell{c}{4\\8\\8 \\ (int)}} & \multirow{5}{*}{12.8} & REMIND(2.0)& 58.51 & 59.94 & 59.87\\
&&&& REMIND(3.0)  & 61.23 & 61.02 & 61.00\\
&&&& Ours(1.0) & 56.40 & 56.98 & 57.01\\
&&&& Ours(2.0) & \textbf{61.27} & \textbf{62.49} & \textbf{62.30}\\
&&&& $-\mathcal{L}_{CE}$ & 56.11 & 60.19 & 60.07 \\

\hline
\end{tabular}
}
\end{center}
\end{table}

\minisection{Many-task evaluation (25/50 steps setting).} Here the number of rehearsed samples is set to $\mathcal{N}=200$ because there are less samples in each step. The results for 25 steps on ImageNet-Subset are shown in Table~\ref{tab:base50tasks5}. We obtain state-of-the-art with 3.12\%, 0.93\% and 4.50\% higher than PODNet, REMIND (block 4.0) and REMIND (block 3.0) respectively. The hardest setting is the 50 steps split, where only 1 class is viewed at every time step. We show our performance in Table~\ref{tab:base50tasks5}. It is 8.49\% higher than GDumb in the offline setting and 1.21\% better than REMIND in the online setting. 

For the many-task evaluation of CIFAR100 shown in Table~\ref{tab:cifar100}, we got marginally better than PODNet in 50 steps and worse in 25 steps under higher memory allocation. With lower memory allocation we still got competitive performances. In conclusion, from the many-task evaluation, we observe that bias correction methods suffer from a drop in performance with more time steps, while our model obtains better results and without much drop in performance when the number of tasks increases.

\minisection{Equal split with 10 tasks (9 steps) on ImageNet-Subset.} In the equal split setting, the 100 classes are divided into 10 tasks with 10 classes each. The top-5 accuracies in each time step are shown in Table~\ref{tab:base10tasks10} (For comparison, top-5 accuracy is adopted here since it is commonly used in this case). For this more challenging setting, REMIND obtains 8.5\% lower results than ours, which is due to the less flexible backbone model. Especially here, it makes more sense to perform replay on a lower layer. Also note that in this setting, we get competitive results compared with the methods BiC~\cite{wu2019large} and RPSnet~\cite{rajasegaran2019random}. However, these methods are offline and perform multiple loops over the data.

\minisection{Equal split with 5 tasks (4 steps) on CIFAR10.}
Several existing online methods cannot be straightforwardly applied to large datasets. To be able to compare to them, we also include results on CIFAR10 in Table~\ref{tab:cifar10}.
We divide the 10 classes into 5 tasks with 2 classes each. On this setting, we got 48.4\%, which is 2.6\% higher than the best reported results of GDumb. We also outperform the REMIND method with more than a 3\% margin.

\begin{table}
\begin{center}
\caption{\label{tab:base10tasks10} Comparison on ImageNet-Subset, we show \alert{top-5} accuracy with 10 classes as the first task and 9 steps each with 10 classes. For REMIND and our method, we show the replay layer (block number) in brackets. The highest numbers in offline and online settings are highlighted.}
\scalebox{0.7}{
\begin{tabular}{|c|c|c|c|c|c|c|c|c|}

\hline

\multicolumn{2}{|c|}{Methods } & iCaRL & RPSnet  & BiC &  \tabincell{c}{REMIND \\ (4.0)} & \tabincell{c}{Ours \\ (3.0)} & \tabincell{c}{Ours \\ (2.0)}& \tabincell{c}{Ours \\ (1.0)}\\
\hline
\multicolumn{2}{|c|}{Online } & \multicolumn{3}{c|}{\xmark} & \multicolumn{4}{c|}{\cmark} \\
\hline
\multirow{3}{*}{\tabincell{c}{Exem.\\ Info}} & \tabincell{c}{Num.} &  \multicolumn{3}{c|}{\tabincell{c}{2000 \\ (20*100)}} &  \multicolumn{4}{c|}{\tabincell{c}{130000 \\ (1300*100)}}\\
\cline{2-9}
& \tabincell{c}{Shape \\ (CHW)} &  \multicolumn{3}{c|}{\tabincell{c}{3*224*224 \\ (integer)}} &  \multicolumn{4}{c|}{\tabincell{c}{32*7*7 \\ (integer)}}\\
\cline{2-9}
& \tabincell{c}{Mem. \\ (MB)}&  \multicolumn{3}{c|}{301.06} &  \multicolumn{4}{c|}{\tabincell{c}{203.84}} \\

\hline

\multirow{11}{*}{Acc.} & 1 & 99.3 & 100 & 98.4 & 98.4 & 98.4 & 98.4 & 98.4\\
& 2 & 97.2 & 97.4 & 96.2 & 91.6 & 93.3 & 93.5 & 94.1 \\
& 3 & 93.5 & 94.3 & 94.0 & 87.1 & 90.5 & 91.1 & 92.7 \\
&4  & 91.0    & 92.7  & 92.9  & 82.2  & 87.2  & 87.7 & 90.2  \\
&5  & 87.5  & 89.4  & 91.1  & 79.7  & 85.3 & 85.5 & 89.2  \\
&6  & 82.1  & 86.6  & 89.4  & 77.7  & 84.0 & 85.0 & 87.8  \\
&7  & 77.1  & 83.9  & 88.1  & 74.8   & 81.0 & 83.7 & 85.7  \\
&8   & 72.8  & 82.4  & 86.5  & 72.8 & 80.9 & 82.7 & 85.4  \\
&9   & 67.1  & 79.4  & 85.4  & 72.2  & 80.8 & 83.4 & 84.7  \\
&10  & 63.5  & 74.1  & \textbf{84.4}  & 70.9  & 79.6 & 81.8 & \textbf{83.9 }  \\
\cline{2-9}
& AOC & 83.1 & 88.0 & \textbf{90.6} & 80.7 & 86.1 & 87.1 & \textbf{89.2}\\

\hline
\end{tabular}}
\end{center}

\end{table}

\begin{table}
\begin{center}
\caption{\label{tab:cifar10}Comparison on CIFAR10 dataset, we show the LAST accuracy with 2 classes as the first task and 4 steps each with 2 classes. For REMIND and our method, we show the replay layer (block number) in brackets. The highest numbers in each row are highlighted.}

\scalebox{0.85}{
\begin{tabular}{|c|c|c|c|c|c|}
\hline
\multirow{2}{*}{Online} & \multicolumn{3}{c|}{Exemplar info} & \multirow{2}{*}{Methods} & LAST \\
\cline{2-4}
\cline{6-6}
 & Num.  & \tabincell{c}{Shape \\ (C*H*W)} & \tabincell{c}{Mem. \\ (MB)}  & & 4 steps \\

\hline
\multirow{11}{*}{\cmark } & \multirow{11}{*}{500} & \multirow{11}{*}{\tabincell{c}{3*32*32 \\ (integer)}} & \multirow{11}{*}{1.536} & Finetuning  & 18.5 \\
&&&& AGEM & 18.5\\
&&&& BGD& 18.2 \\
&&&& GEM & 20.1 \\
&&&& GSS-Greedy & 28.0 \\
&&&& HAL  & 32.1 \\
&&&& ER & 33.3 \\
&&&& MIR  & 34.5\\
&&&& GMED(ER)  & 35.0\\
&&&& GMED(MIR)   & 35.5\\
&&&& GDumb  & 45.8\\
\hline

\multirow{2}{*}{\cmark} & \multirow{2}{*}{24000} &  \multirow{2}{*}{\tabincell{c}{1*8*8 \\ (integer)}} & \multirow{2}{*}{1.536} & REMIND(3.0)  & 45.2\\

&&&& Ours(2.0) & \textbf{48.4}\\

\hline
\end{tabular}
}
\end{center}
\end{table}

\begin{figure*}[t]
\begin{minipage}[b]{0.33\linewidth}
\centering
\includegraphics[width=\textwidth]{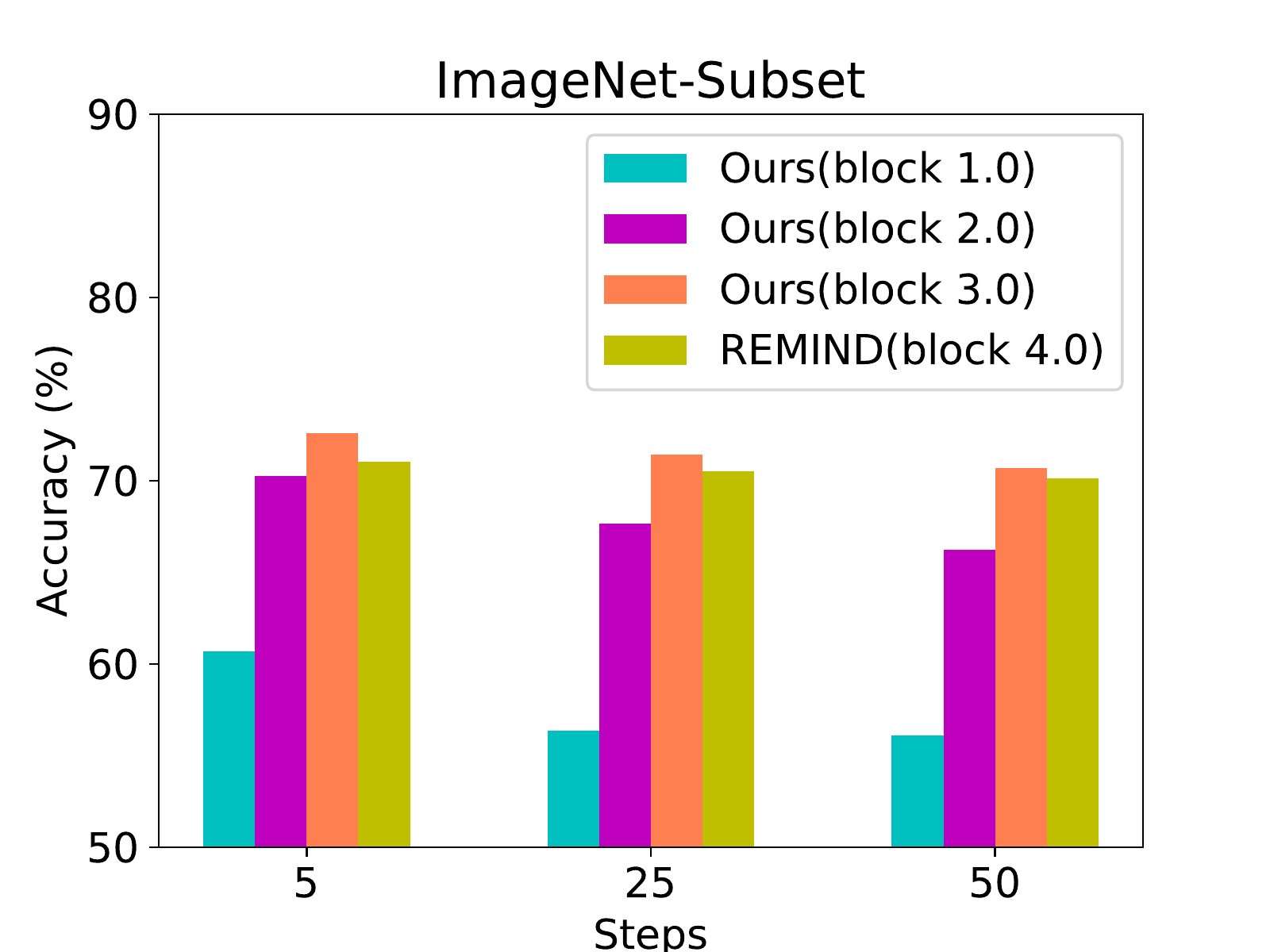}
\subcaption{{\textit{1st} task is 50 classes then 5/25/50 steps}}
 \label{fig:imagenet_subset_ablate_block_n}
\end{minipage}
\begin{minipage}[b]{0.33\linewidth}
\centering
\includegraphics[width=\textwidth]{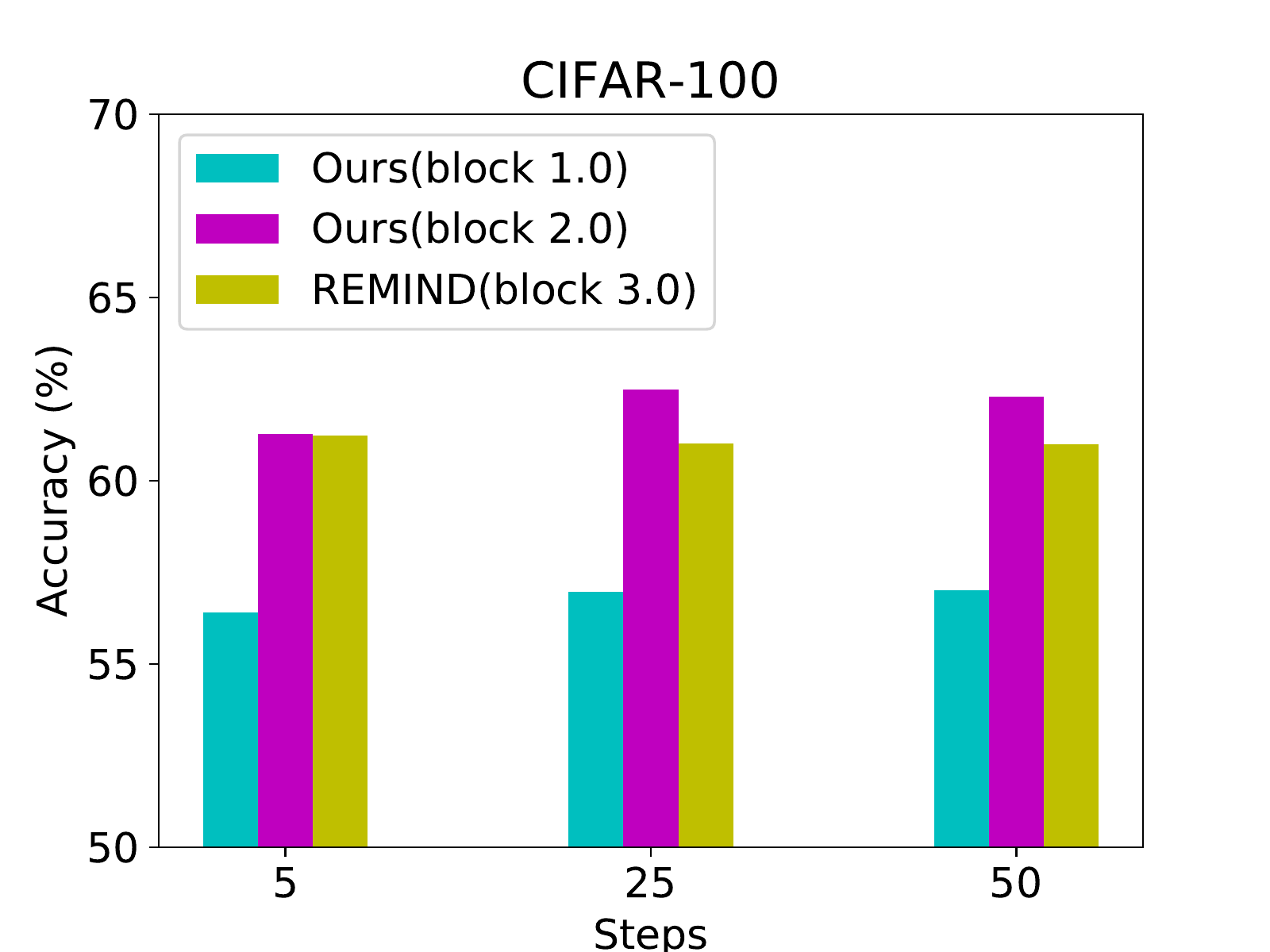}
\subcaption{\alert{\textit{1st} task is 50 classes then 5/25/50 steps}}
 \label{fig:cifar_ablate_block_n}
\end{minipage}
\begin{minipage}[b]{0.3\linewidth}
\centering
\includegraphics[width=\textwidth]{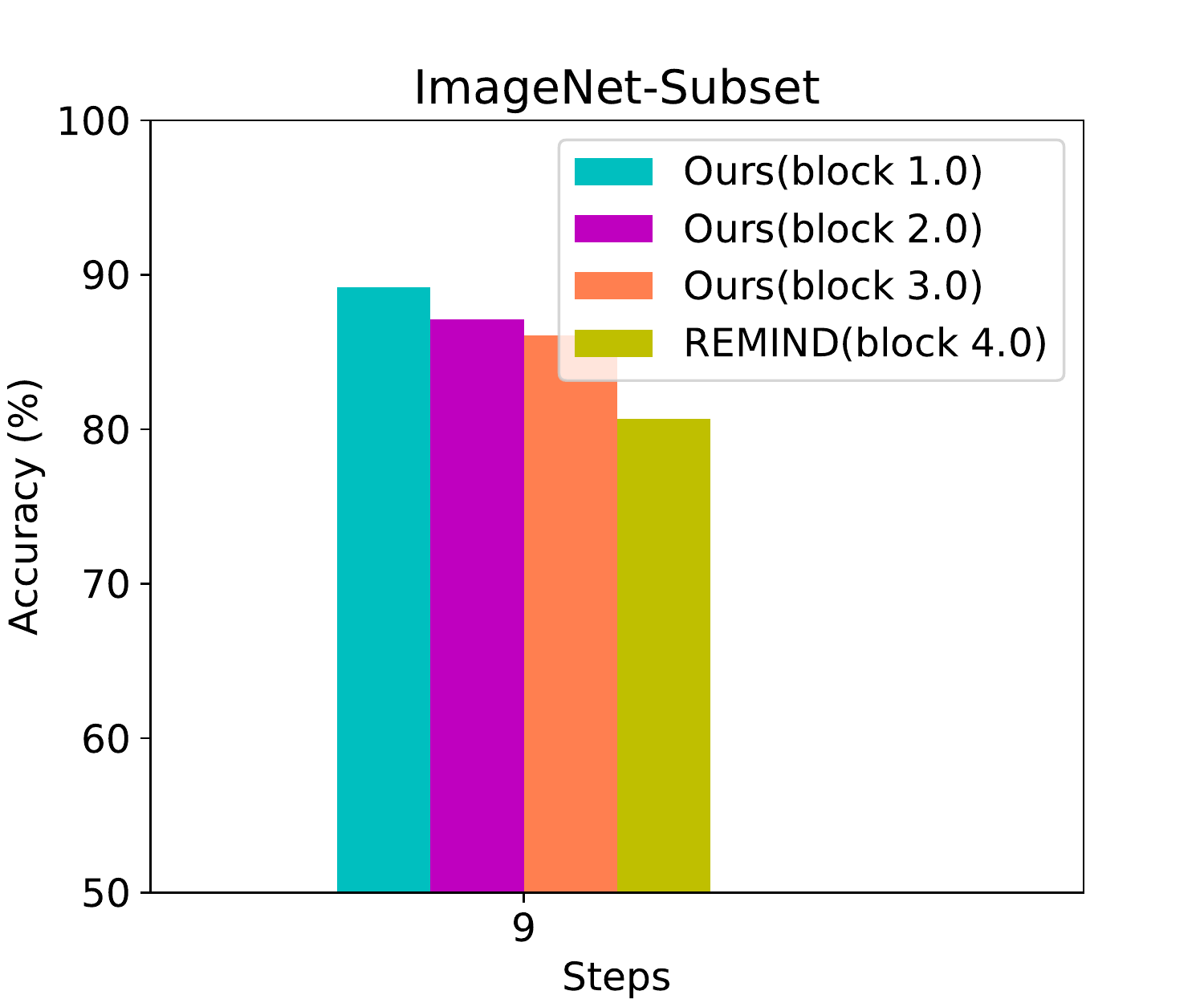}
\subcaption{{\textit{1st} task is 10 classes then 9 steps}}
\label{fig:imagenet_ablate_block_n_10by10}
\end{minipage}

\caption{
\alert{
Ablation study on the block number $n$ on ImageNet-Subset and CIFAR100 with various settings. The backbone for ImageNet-Subset is a 4-block Resnet-18 and for CIFAR100 is a 3-block Resnet-32. We show top-1 accuracy in (a) and (b), and top-5 accuracy (c). 
}
}
\label{fig:ablate_block_n}
\end{figure*}

\subsection{Ablation study}

One of the key ingredients of the ACAE-REMIND method is the auxiliary classification loss that is used during the training of the auto-encoder (see Eq.~\ref{eq:acae_loss}). 
This loss ensures that the compression does not remove the features that are crucial for classification. Here we ablate this factor. To show the impact of the classification loss, we evaluate the classification accuracy after compression with and without the loss (directly after Step 2), and compare this to the results that would be obtained with the uncompressed features (see Table~\ref{tab:ablation_cls}). The results show that classification loss mitigates the classification drop that occurs due to compression. Finally, we have also ablated the loss in  Table~\ref{tab:base50tasks5} and Table~\ref{tab:cifar100} on our best performing setting (indicated by rows with $-\mathcal{L}_{CE}$). The results show that the loss does greatly improve results resulting in a performance gain of 2-5\%.

\begin{table}
\begin{center}
\caption{\label{tab:ablation_cls}Ablation study of classification loss on CIFAR100 and ImageNet-Subset. The features are replayed from block 2.0 for CIFAR100 and block 1.0/2.0/3.0 for ImageNet-Subset.}

\scalebox{0.75}{
\begin{tabular}{|c|c|c|c|c|}
\hline

\multirow{2}{*}{method name}  & CIFAR100 &  \multicolumn{3}{c|}{ImageNet-Subset} \\
\cline{2-5}

& block 2.0 & block 3.0 & block 2.0 & block 1.0 \\
\hline\hline
Uncompressed features & 76.00\% & 80.96\% & 80.96\%& 80.96\%\\
\hline
\tabincell{c}{ACAE replay(w/o $\mathcal{L}_{CE}$)} & 73.31\%  & 78.80\% & 77.64\% & 76.52\%\\
\hline
\tabincell{c}{ACAE replay(w/ $\mathcal{L}_{CE}$)} & 74.45\%  & 79.40\% & 79.76\% &80.44\%\\
\hline
\end{tabular}
}
\end{center}
\end{table}

\alert{To better evaluate the influence of the block number $n$ (the block where we introduce the ACAE-REMIND as seen in Fig.~\ref{fig:CL}) we have included Fig.~\ref{fig:ablate_block_n}.
Here we show a comparison on ImageNet-Subset and CIFAR-100 under the 5, 25 and 50 steps settings. We can conclude that under these settings, the performances are increasing with the feature replay from the first block to the penultimate block, and then decreasing when replaying on the last block. As can be seen the optimal block is relatively stable with respect to the number of steps while keeping the same amount of data for the first task. If we however reduce the number of data for the first task, it becomes more difficult to learn a good backbone network and, as expected, the optimal $n$ decreases: in Fig.~\ref{fig:ablate_block_n}(c) we can see that $n=1$ yields optimal results.

}

\section{Conclusions}

In this paper, we proposed an extension to the REMIND method, called ACAE-REMIND. We propose a stronger compression module based on an auxiliary classifier auto-encoder that allows to move the feature replay to lower layers. The method is memory efficient and obtains better performance. In evaluation, we perform a comparison over multiple metrics among competitive methods. The strength of our model lies in the fact that with high compression ratio, we could save more feature exemplars than image exemplars. Especially, when the first task is relatively small (the 10-task scenario in ImageNet-Subset and 5-task in CIFAR10) we outperform REMIND with a large margin.
As future work, we are interested in extending this framework to other continual learning problems.

\section*{Acknowledgements}
We acknowledge the support from Huawei Kirin Solution, the Spanish Government funding for projects PID2019-104174GB-I00 and RTI2018-102285-A-I00, and Kai acknowledges the Chinese Scholarship Council (CSC) No.201706170035. Luis acknowledges the Ramón y Cajal fellowship RYC2019-027020-I.

{\small
\bibliographystyle{ieee_fullname}
\bibliography{refs}

\begin{thebibliography}{10}\itemsep=-1pt

\bibitem{aljundi2019online}
Rahaf Aljundi, Eugene Belilovsky, Tinne Tuytelaars, Laurent Charlin, Massimo
  Caccia, Min Lin, and Lucas Page-Caccia.
\newblock Online continual learning with maximal interfered retrieval.
\newblock In {\em Advances in Neural Information Processing Systems}, pages
  11849--11860, 2019.

\bibitem{aljundi2017expert}
Rahaf Aljundi, Punarjay Chakravarty, and Tinne Tuytelaars.
\newblock Expert gate: Lifelong learning with a network of experts.
\newblock In {\em Proceedings of the IEEE Conference on Computer Vision and
  Pattern Recognition}, pages 3366--3375, 2017.

\bibitem{aljundi2019gradient}
Rahaf Aljundi, Min Lin, Baptiste Goujaud, and Yoshua Bengio.
\newblock Gradient based sample selection for online continual learning.
\newblock In {\em Advances in Neural Information Processing Systems}, pages
  11816--11825, 2019.

\bibitem{chaudhry2020using}
Arslan Chaudhry, Albert Gordo, Puneet~K Dokania, Philip Torr, and David
  Lopez-Paz.
\newblock Using hindsight to anchor past knowledge in continual learning.
\newblock {\em arXiv preprint arXiv:2002.08165}, 2020.

\bibitem{chaudhry2018efficient}
Arslan Chaudhry, Marc'Aurelio Ranzato, Marcus Rohrbach, and Mohamed Elhoseiny.
\newblock Efficient lifelong learning with a-gem.
\newblock {\em arXiv preprint arXiv:1812.00420}, 2018.

\bibitem{de2019continual}
Matthias De~Lange, Rahaf Aljundi, Marc Masana, Sarah Parisot, Xu Jia,
  Ale{\v{s}} Leonardis, Gregory Slabaugh, and Tinne Tuytelaars.
\newblock A continual learning survey: Defying forgetting in classification
  tasks.
\newblock {\em arXiv preprint arXiv:1909.08383}, 2019.

\bibitem{imagenet_cvpr09}
Jia Deng, Wei Dong, Richard Socher, Li-Jia Li, Kai Li, and Li Fei-Fei.
\newblock Imagenet: A large-scale hierarchical image database.
\newblock In {\em 2009 IEEE conference on computer vision and pattern
  recognition}, pages 248--255. Ieee, 2009.

\bibitem{douillard2020podnet}
Arthur Douillard, Matthieu Cord, Charles Ollion, Thomas Robert, and Eduardo
  Valle.
\newblock Podnet: Pooled outputs distillation for small-tasks incremental
  learning.
\newblock In {\em Proceedings of the IEEE European Conference on Computer
  Vision (ECCV)}, volume~2, page~6. Springer, 2020.

\bibitem{farquhar2018towards}
Sebastian Farquhar and Yarin Gal.
\newblock Towards robust evaluations of continual learning.
\newblock {\em arXiv preprint arXiv:1805.09733}, 2018.

\bibitem{hayes2019remind}
Tyler~L Hayes, Kushal Kafle, Robik Shrestha, Manoj Acharya, and Christopher
  Kanan.
\newblock Remind your neural network to prevent catastrophic forgetting.
\newblock {\em arXiv preprint arXiv:1910.02509}, 2019.

\bibitem{hou2019learning}
Saihui Hou, Xinyu Pan, Chen~Change Loy, Zilei Wang, and Dahua Lin.
\newblock Learning a unified classifier incrementally via rebalancing.
\newblock In {\em Proceedings of the IEEE Conference on Computer Vision and
  Pattern Recognition}, pages 831--839, 2019.

\bibitem{iscen2020memory}
Ahmet Iscen, Jeffrey Zhang, Svetlana Lazebnik, and Cordelia Schmid.
\newblock Memory-efficient incremental learning through feature adaptation.
\newblock {\em arXiv preprint arXiv:2004.00713}, 2020.

\bibitem{jegou2010product}
Herve Jegou, Matthijs Douze, and Cordelia Schmid.
\newblock Product quantization for nearest neighbor search.
\newblock {\em IEEE transactions on pattern analysis and machine intelligence},
  33(1):117--128, 2010.

\bibitem{jin2020gradient}
Xisen Jin, Junyi Du, and Xiang Ren.
\newblock Gradient based memory editing for task-free continual learning.
\newblock {\em arXiv preprint arXiv:2006.15294}, 2020.

\bibitem{JDH17}
Jeff Johnson, Matthijs Douze, and Herv{\'e} J{\'e}gou.
\newblock Billion-scale similarity search with gpus.
\newblock {\em arXiv preprint arXiv:1702.08734}, 2017.

\bibitem{kingma2014adam}
Diederik~P Kingma and Jimmy Ba.
\newblock Adam: A method for stochastic optimization.
\newblock {\em arXiv preprint arXiv:1412.6980}, 2014.

\bibitem{kirkpatrick2017overcoming}
James Kirkpatrick, Razvan Pascanu, Neil Rabinowitz, Joel Veness, Guillaume
  Desjardins, Andrei~A Rusu, Kieran Milan, John Quan, Tiago Ramalho, Agnieszka
  Grabska-Barwinska, et~al.
\newblock Overcoming catastrophic forgetting in neural networks.
\newblock {\em Proceedings of the national academy of sciences},
  114(13):3521--3526, 2017.

\bibitem{kramer1991nonlinear}
Mark~A Kramer.
\newblock Nonlinear principal component analysis using autoassociative neural
  networks.
\newblock {\em AIChE journal}, 37(2):233--243, 1991.

\bibitem{krizhevsky2009learning}
Alex Krizhevsky, Geoffrey Hinton, et~al.
\newblock Learning multiple layers of features from tiny images.
\newblock 2009.

\bibitem{li2017learning}
Zhizhong Li and Derek Hoiem.
\newblock Learning without forgetting.
\newblock {\em IEEE transactions on pattern analysis and machine intelligence},
  40(12):2935--2947, 2017.

\bibitem{liu2020generative}
Xialei Liu, Chenshen Wu, Mikel Menta, Luis Herranz, Bogdan Raducanu, Andrew~D
  Bagdanov, Shangling Jui, and Joost van~de Weijer.
\newblock Generative feature replay for class-incremental learning.
\newblock In {\em Proceedings of the IEEE/CVF Conference on Computer Vision and
  Pattern Recognition Workshops}, pages 226--227, 2020.

\bibitem{lopez2017gradient}
David Lopez-Paz and Marc'Aurelio Ranzato.
\newblock Gradient episodic memory for continual learning.
\newblock In {\em Advances in neural information processing systems}, 2017.

\bibitem{mallya2018piggyback}
Arun Mallya, Dillon Davis, and Svetlana Lazebnik.
\newblock Piggyback: Adapting a single network to multiple tasks by learning to
  mask weights.
\newblock In {\em Proceedings of the European Conference on Computer Vision
  (ECCV)}, pages 67--82, 2018.

\bibitem{mallya2018packnet}
Arun Mallya and Svetlana Lazebnik.
\newblock Packnet: Adding multiple tasks to a single network by iterative
  pruning.
\newblock In {\em Proceedings of the IEEE Conference on Computer Vision and
  Pattern Recognition}, pages 7765--7773, 2018.

\bibitem{masana2020class}
Marc Masana, Xialei Liu, Bartlomiej Twardowski, Mikel Menta, Andrew~D Bagdanov,
  and Joost van~de Weijer.
\newblock Class-incremental learning: survey and performance evaluation.
\newblock {\em arXiv preprint arXiv:2010.15277}, 2020.

\bibitem{masana2020ternary}
Marc Masana, Tinne Tuytelaars, and Joost van~de Weijer.
\newblock Ternary feature masks: continual learning without any forgetting.
\newblock {\em arXiv preprint:2001.08714}, 2020.

\bibitem{mccloskey1989catastrophic}
Michael McCloskey and Neal~J Cohen.
\newblock Catastrophic interference in connectionist networks: The sequential
  learning problem.
\newblock In {\em Psychology of learning and motivation}, volume~24, pages
  109--165. Elsevier, 1989.

\bibitem{prabhu2020gdumb}
Ameya Prabhu, Philip~HS Torr, and Puneet~K Dokania.
\newblock Gdumb: A simple approach that questions our progress in continual
  learning.
\newblock In {\em European Conference on Computer Vision}, pages 524--540.
  Springer, 2020.

\bibitem{rajasegaran2019random}
Jathushan Rajasegaran, Munawar Hayat, Salman~H Khan, Fahad~Shahbaz Khan, and
  Ling Shao.
\newblock Random path selection for continual learning.
\newblock In {\em Advances in Neural Information Processing Systems}, pages
  12669--12679, 2019.

\bibitem{rebuffi2017icarl}
Sylvestre-Alvise Rebuffi, Alexander Kolesnikov, Georg Sperl, and Christoph~H
  Lampert.
\newblock icarl: Incremental classifier and representation learning.
\newblock In {\em Proceedings of the IEEE conference on Computer Vision and
  Pattern Recognition}, pages 2001--2010, 2017.

\bibitem{rolnick2019experience}
David Rolnick, Arun Ahuja, Jonathan Schwarz, Timothy Lillicrap, and Gregory
  Wayne.
\newblock Experience replay for continual learning.
\newblock In {\em Advances in Neural Information Processing Systems}, pages
  350--360, 2019.

\bibitem{serra2018overcoming}
Joan Serra, Didac Suris, Marius Miron, and Alexandros Karatzoglou.
\newblock Overcoming catastrophic forgetting with hard attention to the task.
\newblock {\em arXiv preprint arXiv:1801.01423}, 2018.

\bibitem{shin2017continual}
Hanul Shin, Jung~Kwon Lee, Jaehong Kim, and Jiwon Kim.
\newblock Continual learning with deep generative replay.
\newblock In {\em Advances in Neural Information Processing Systems}, 2017.

\bibitem{wu2018memory}
Chenshen Wu, Luis Herranz, Xialei Liu, Joost van~de Weijer, Bogdan Raducanu,
  et~al.
\newblock Memory replay gans: Learning to generate new categories without
  forgetting.
\newblock In {\em Advances in Neural Information Processing Systems}, pages
  5962--5972, 2018.

\bibitem{wu2019large}
Yue Wu, Yinpeng Chen, Lijuan Wang, Yuancheng Ye, Zicheng Liu, Yandong Guo, and
  Yun Fu.
\newblock Large scale incremental learning.
\newblock In {\em Proceedings of the IEEE Conference on Computer Vision and
  Pattern Recognition}, pages 374--382, 2019.

\bibitem{zeno2018task}
Chen Zeno, Itay Golan, Elad Hoffer, and Daniel Soudry.
\newblock Task agnostic continual learning using online variational bayes.
\newblock {\em arXiv preprint arXiv:1803.10123}, 2018.

\end{thebibliography}
}

\end{document}